\documentclass{article}

     \usepackage[final]{drlw_neurips_2019}

\usepackage[utf8]{inputenc} % allow utf-8 input
\usepackage[T1]{fontenc}    % use 8-bit T1 fonts
\usepackage{hyperref}       % hyperlinks
\usepackage{url}            % simple URL typesetting
\usepackage{booktabs}       % professional-quality tables
\usepackage{amsfonts}       % blackboard math symbols
\usepackage{nicefrac}       % compact symbols for 1/2, etc.
\usepackage{microtype}      % microtypography

\usepackage{array}
\usepackage{tabularx}
\usepackage{esvect}
\usepackage{pgfplots}
\usepackage{algpseudocode}
\usepackage{mathtools}
\usepackage{multicol}
\usepackage{amssymb}
\usepackage{amsmath}
\usepackage{listings}

\usepackage{wrapfig}
\usepackage{graphicx}
\usepackage{caption}
\usepackage{subcaption}

\usepackage[linesnumbered,ruled]{algorithm2e}
\makeatletter
\algdef{SE}[SUBALG]{Indent}{EndIndent}{}{\algorithmicend\ }%
\algtext*{Indent}
\algtext*{EndIndent}

\DeclareMathOperator*{\argmin}{arg\,min}
\DeclareMathOperator{\E}{\mathbb{E}}

\newcommand{\context}{\textbf{c}}
\newcommand{\pic}{\pi_{\context}}
\newcommand{\ci}{c^{(i)}}
\newcommand{\pici}{\pi_{\ci}}
\newcommand{\pim}{\pi_o}

\newcommand{\defeq}{\vcentcolon=}

\title{Preventing Imitation Learning with Adversarial Policy Ensembles
}

\author{
  Albert Zhan\\
  UC Berkeley\\
  \texttt{albertzhan@berkeley.edu}\\
  \And
  Stas Tiomkin\\
  UC Berkeley\\
  \texttt{stas@berkeley.edu}\\
  \And
  Pieter Abbeel\\
  UC Berkeley\\
  \texttt{pabbeel@cs.berkeley.edu}
}

\begin{document}

\maketitle

\begin{abstract}
    Imitation learning can reproduce policies by observing experts, which poses a problem regarding policy privacy. 
    Policies, such as human, or policies on deployed robots, can all be cloned without consent from the owners. 
    How can we protect against external observers cloning our proprietary policies? 
    To answer this question we introduce a new reinforcement learning framework, where we train an ensemble of near-optimal policies, whose demonstrations are guaranteed to be useless for an external observer. 
    We formulate this idea by a constrained optimization problem, where the objective is to improve proprietary policies, and at the same time deteriorate the virtual policy of an eventual external observer. 
    We design a tractable algorithm to solve this new optimization problem by modifying the standard policy gradient algorithm. 
    Our formulation can be interpreted in lenses of confidentiality and adversarial behaviour, which enables a broader perspective of this work. We demonstrate the existence of  ``non-clonable'' ensembles, providing a solution to the above optimization problem, which is calculated by our modified policy gradient algorithm. 
    To our knowledge, this is the first work regarding the protection of policies in Reinforcement Learning.
\end{abstract}

\section{Introduction}

\begin{figure}
          \includegraphics[scale=0.04]{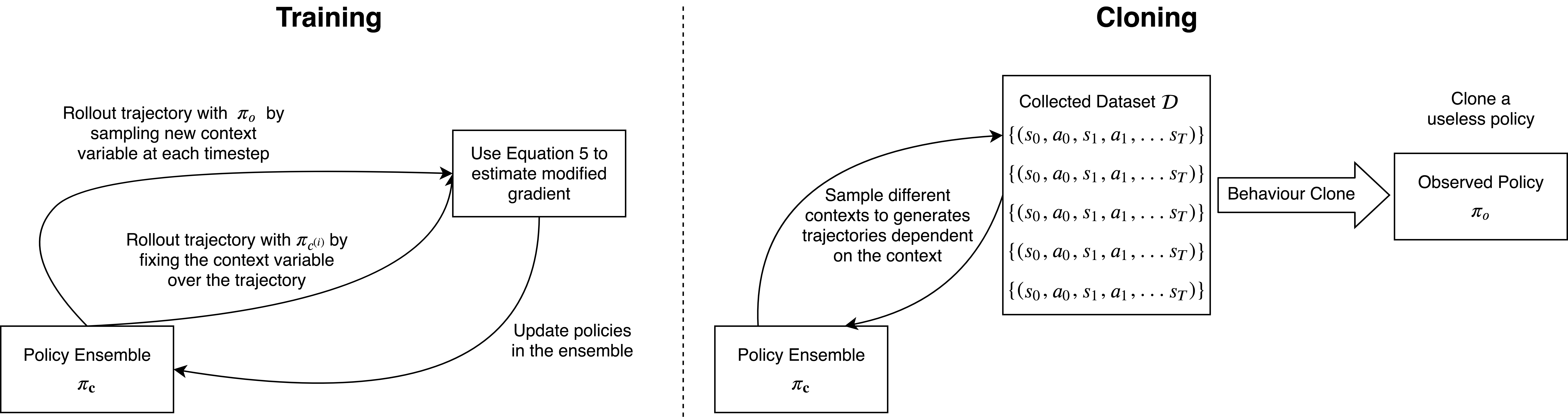}
          \captionof{figure}{\textbf{Confidentiality scheme}: \textbf{Left} During training, optimize a Policy Ensemble by estimating gradients using both the policies in the ensemble and the fictitious observer policy. \textbf{Right} When collecting a dataset for cloning, the context variable is marginalized out. Thus cloning the Policy Ensemble can result in a useless policy}
          \label{fig:scheme}
    \end{figure}

    Imitation learning and behavioral cloning provide really strong ability to create powerful policies, as seen in robotic tasks (\cite{DART, oneshotIL, BCLimitations, end2endCondIL, alvinn, end2endSelfDriving}).
    Other fields in machine learning have developed methods to ensure privacy
    (\citet{privateML, PATE}),
    however, none have examined \textit{protection} against policy cloning. 
    In this work, we tackle the issue of protecting policies by training policies that aim to prevent an external observer from using behaviour cloning. 
    Our approach draws inspiration from imitating human experts, 
    who can near-optimally accomplish given tasks.
    The setting which we analyze is presented in Figure \ref{fig:scheme}.
    We wish to find a collection of experts, which as an ensemble can perform a given task well, however, also targets behaviour cloning through adversarial behaviour. 
    Another interpretation is that this collection of experts represents the worst case scenario for behaviour cloning on how to perform a task "good enough". 

    Imitation learning frameworks generally make certain assumptions of the optimality of the demonstrations (\cite{maxentIRL, controlasoptimalinference}), yet never considered the scenario when the experts specifically attempt to be adversarial to the imitator.
    We pose the novel question regarding this assumption:
    does there exist a set of experts that are adversarial to an external observer trying to behaviour clone?    

   We propose Adversarial Policy Ensembles (APE), a method
    that simultaneously optimizes the performance of the ensemble and minimizes the performance of policies eventually obtained
    from cloning it. 
    Our experiments show that
    APE do not suffer much performance loss from an optimal policy,
    while causing, on average, the cloned policy to experience over $5$ times degradation compared to the optimal policy.

    Our main contributions can be summarized as follows:
    \begin{itemize}
        \item We introduce a novel method APE, as well as the mathematical justification of the notion of adversarial experts.
        \item By modifying Policy Gradient (\citet{PG}), a common reinforcement learning algorithm, we suggest a tractable scheme for finding an optimal solution for this objective.
        \item We demonstrate the solution by numerical simulations, where we show that a cloned policy is crippled even after collecting a significantly large number of samples from a policy ensemble. 
    \end{itemize}

    To our knowledge, not only is this the first work regarding the protection of policies in reinforcement learning, but it is also the first to represent adversarial experts.

\section{Preliminaries}
We develop APE in the standard framework of Reinforcement Learning (RL). The main components we use are Markov Decision Processes, Policy Gradient (\citet{PG}), policy ensembles, and behaviour cloning, which we review below.
\subsection{Markov Decision Process}
A discrete-time finite-horizon discounted Markov decision process (MDP) $\mathcal{M}$ is defined by 
$(\mathcal{S}, \mathcal{A}, r, p, p_0, \gamma, T)$ where 
$\mathcal{S}$ is the state space, 
$\mathcal{A}$ is the action space, 
$r : \mathcal{S} \times \mathcal{A} \rightarrow \mathbb{R}$ is the reward function, 
$p(s_{t+1} | s_t, a_t)$ is the transition probability distribution,  
$p_0 : \mathcal{S} \rightarrow \mathbb{R}^{+}$ is the initial state distribution, 
$\gamma \in (0, 1)$ is the discount factor,
and $T$ is the time horizon. 
A trajectory $\tau \sim \rho_\pi$, sampled from 
$p$ and a 
policy $\pi : \mathcal{S} \times \mathcal{A} \rightarrow \mathbb{R}^+$, is defined to be the states and actions tuple $(s_0, a_0, ... s_{T-1}, a_{T-1}, s_T)$, whose distribution is characterized by $\rho_\pi$. 
Define the return of a trajectory to be
$r(\tau) = \sum_{t=0}^{T-1} \gamma ^{t} r(s_{t}, a_{t})$ 
to be the sum of discounted rewards seen along the trajectory, 
and define a value function $V^\pi : \mathcal{S} \rightarrow \mathbb{R}$ to be expected return of a trajectory starting from state $s$, under the policy $\pi$.
The goal of reinforcement learning is to find a policy 
that maximizes the expected return $\E_{\tau \sim \rho_\pi} [r(\tau)]$. 

\subsection{Policy Gradient}

Policy Gradient (PG) \citet{PG} 
aim to directly learn the optimal policy $\pi$, parameterized by $\theta$,
by repeatedly estimating the gradient of the expected return, in one of many forms, shown in \citet{gae}.
In our work, we follow notation similar to that of \citet{gae, PPO} and estimate $\nabla_\theta \E_{\tau \sim \rho_\pi}[r(\tau)]$ using the advantage, which is estimated from a trajectory $\tau$, 
$ A^\pi_\tau (t) = R_\tau (t) - V^\pi (s_t)$, where 
% $R(t) = \sum_{t' = t}^{T} \gamma^{t' - t} r(s_{t'}, a_{t'})$ 
$R_\tau (t) = \sum_{t'=t}^{T-1} \gamma ^{t'} r(s_{t'}, a_{t'})$ 
is the sum of the reward following action $a_t$.

Here, the value function is learned simultaneously with the the policy, and so the advantage will use $\hat{V}^\pi$ as an estimate for $V^\pi$.

\subsection{Policy Ensemble (PE)}

We denote a PE by $\pic$, where each $\pici, i \in \{ 1, 2, ... n \}$ represents an expert. 
To rollout the PE, an expert is chosen at random (in our case uniform), and the expert completes a trajectory.
Each expert policy $\pici (a | s)$ can be viewed as a policy conditioned on a latent variable $c$, $\pi(a | s, c)$.

Although $\pic$ consists of multiple policies, it is important to note that it itself is still a policy.

\subsection{Behaviour Cloning}

To behaviour clone an expert policy (\citet{ILOriginal}), a dataset of trajectories $\mathcal{D}$ consisting of state action pairs $(s, a)$ are collected from the the expert rollouts. 
Then, a policy parametized by $\phi$ is trained by maximizing the likelihood of an action given a state, $ \sum_{(s, a) \in \mathcal{D}} \log \pi_\phi (a \mid s)$. 

When cloning $\pic$, $\mathcal{D}$ will not contain information of the latent variable $c$, and so the \textit{cloned} policy will marginalize it out. Thus, the observer will clone:

\begin{align}
\label{eqn:observe}
\pim ({a} \mid {s}) 
\defeq
\sum_i p(\ci \mid s ) \pici ( a \mid s)
\end{align}

We stress that this policy does not exist until $\pic$ is behaviour cloned. 
$\pim$ is a fictitious policy to represent what would happen in the \textit{best} case scenario of the observer having access to \textit{infinite} data from $\pic$ to clone into $\pim$.

    The scope of this paper is to specifically prevent behavioral cloning from succeeding. Other imitation learning approaches such as inverse reinforcement learning  (\citet{IRL, algoIRL, nonlinear}) and adversarial imitation learning 
    (\citet{GAIL, VDB})
    require rollouts of non-expert policies in the environment, which may be costly, and thus are not considered.

\section{Related Work}
\label{sec:related}
    
    \textbf{Adversarial Attacks in RL:}
     Our notion of adversarial policies is inextricably related to other adversarial methods that target RL such as \citet{DRLAttack}, and \citet{vulnerablePolicies}, that add adversarial perturbations to policy input during training. Other adversarial attacks include poisoning the batch of data used when training RL (\citet{poison}), and exploitation in the multi-agent setting (\citet{gleaveAttackRL}). 
     However, these methods all present as active attacks for various learning techniques. 
     Our method, instead, passively protects against cloning. 
     
     \textbf{Privacy in RL:} With regards to protection, our work is related to differential privacy (\citet{privateML}).
     Differential privacy in RL can be used to create private Q-functions
     (\citet{privateQ}) or private policies (\citet{privatePolicy}), which have private reward functions or private policy evaluation.
     However, we would like to emphasize that our motivation is to prevent cloning, and thus protecting the policies, rather than protecting against differentiating between reward functions and policies. 

    \textbf{Imitation Learning:} 
    Since we comply to the standard imitation learning setting of cloning from a dataset with many experts providing the demonstrations, latent variables w.r.t. imitation learning is well-studied. 
    For example,
    \citet{end2endCondIL} show that conditioning on context representation can make imitation learning a viable option for autonomous driving. 
    \citet{infogail} demonstrate that the latent contextual information in expert trajectories is often semantically meaningful. 
    As well,
    providing extra context variables to condition on also appears in forms of extra queries or providing labels (\cite{RiskAwareIRL, causal_imitation_learning, latentFromDemo}). 
    Our method is different, as we use context variables to prevent imitation learning while learning the policies from scratch, rather than assuming using context variables to increase performance of imitation learning.
    
    \textbf{Multiple Policies:} VALOR, DIAYN, and DADS (\citet{valor, diayn, DADS}) have similar schemes of sampling a latent variable and fixing it throughout a trajectory, although their latent variables (contexts or skills) are used to solve semantically different tasks.
    The reason to solve \textit{different} tasks is due to the objective of using the context variable/skills for learning in an unsupervised setting.
    Our approach differs in both motivation and implementation, as we learn experts that all solve the same task, and constrain so that observers can not clone the policy.
    
    A PE $\pic$ can also be viewed as a Mixture of Experts (\citet{mixexp}), except the gating network assigns probability $1$ to the same expert for an entire trajectory. 
    As such, we do not learn the gating network, although it may still be useful to see $\pic$ as a special case of a mixture of experts where the gating network learns immediately to fix the expert for each trajectory. 
    There are also methods such as OptionGAN (\citet{optiongan}), which uses a mixture of experts model to learn multiple policies as options with access to only expert states.
    
    \citet{novel} also proposes a method to train multiple policies that complete the same task 
    but uses the uncertainty of an autoencoder as a reward augment.
    Their motivation is to find multiple novel policies, while 
    our motivation has no connection to novelty. 
    Due to these differences in motivation, they train each policy one after the other, while our policies are trained simultaneously.

    Policy ensembles are also used in the multi-task and goal conditioned settings in which case the task that is meant to be solved can be viewed as the context.
    Marginalizing out the context variable (Equation \ref{eqn:observe}) of these context-conditioned policies is studied in the case of introducing a KL divergence regularizing term for learning new tasks (\citet{InfoBot}) and for sharing/hiding goals (\citet{LSH}).
    However, the main motivation is different in that both \citet{InfoBot} and \citet{LSH} use $\pim$ to optimize mutual information, while we directly optimize its performance.

\section{Method}

\subsection{Objective}
% We wish to restrict the expected return of the observed policy as shown in \ref{eqn:observe}, while still maximizing the returns of our policy ensemble. We thus modify the standard RL objective to be:

We wish to have experts that can perform the task, while minimizing the possible returns of the cloned policy, denoted in Equation \ref{eqn:observe}.
We modify the standard RL objective to be: 
\begin{gather}
\label{eqn:constrained}
    \argmin_\theta \E_{\tau \sim \rho_{\pim}}[r(\tau)]   
    \text{~~~s.t.~}
    \E_{\tau \sim \rho_{\pic}}[r(\tau)]  \geq \alpha
    % \E_{\tau \sim \rho_{\pim}}[r(\tau)] \leq \alpha
\end{gather}

where $\alpha$ is a parameter that lower bounds the reward of the policy ensemble. 
This translates to maximizing the unconstrained Lagrangian:
\begin{gather}
\label{eqn:unconstrained}
    J(\theta) =  
\E_{\tau \sim \rho_{\pic}} 
[r(\tau)] 
- 
\beta \E_{\tau \sim \rho_{\pim}} 
[ r(\tau)]
\end{gather}

where $1/ \beta$ is the corresponding Lagrangian multiplier, and is subsumed into the returns collected by the policy ensemble. 
We refer to PE that optimizes this objective as Adversarial Policy Ensembles (APE).
% \AZ{begin option 1}
There is a natural interpretation of the objective in Equation \ref{eqn:constrained}. 
Human experts tend to be "good enough", which is reflected in the constraint. 
The minimization is simply finding the most adversarial experts.

% \AZ{Not sure to use prev interpretation of (2) or this one.}

% The constraint tells us that on average, the policies that you execute will be "good enough", while the minimization tells us that the policies should be as adversarial as with respect to the observer. \AZ{end option 2}

Although we assume that the observer can only map states to actions, it may be the case that they can train a sequential policy, which is dependent on its previous states and actions.
% \AZ{very briefly explain what the implication of a sequential policy is} 
Our method can be generalized to sequential policies as well, and the impact of such observers is discussed in the Section \ref{sec:dicussion}.

\subsection{Modified Policy Gradient Algorithm}

Intuitively, since there are the returns of two policies that are being optimized, both should be sampled from to estimate the returns.

We show how we can modify PG to train APE, by 
maximizing Equation \ref{eqn:unconstrained}. 
The two terms suggest a simple scheme to estimate the returns of the policy ensemble twice: once using $\pic$ that we wish to maximize, and a second time using $\pim$, which approximates the returns of an eventual observer who tries to clone the policy ensemble. 
Along with our PE, we train value functions $\Tilde{V}^{\pi_{c^{(i)}}}$ for each expert, jointly parameterized by $\phi$
which estimates $V^{\pici} - \beta V^{\pim}$. The loss function for the value functions of two sampled trajectories $\tau_1, \tau_2$ is 

\begin{gather}
\label{sec: returns}
    J_{\tau_1, \tau_2} (\phi) = 
    % \left[ 
        \sum_{t=0}^{T_1-1} \frac{1}{2} \left(\Tilde{V}_\phi ^{\pici} (s_{t_1}) - R_{\tau_1}(t) \right) ^2
            +
             \sum_{t=0}^{T_2-1} \frac{1}{2} \left(\Tilde{V}_\phi ^ {\pici} (s_{t _2}) + \beta R_{\tau_2}(t) \right)^2
    % \right]
\end{gather}

The policy gradient update from $N_1$ and $N_2$ trajectories is then
\begin{gather}
\label{sec: pgupdate}
    \nabla_\theta J_{{\tau}_1 , {\tau}_2} (\theta)
    \approx
    G_1 + G_2
\end{gather}
where
\begin{gather}
\label{sec: g1}
G_1 = 
    \frac{1}{N_1} 
    \displaystyle\sum_{j=1}^{N_1} 
        \displaystyle\sum_{t=0}^{T_1}
        \nabla_\theta \log{\pici (a_{t1}^{(j)} \mid s_{t1}^{(j)}) 
            \Tilde{A}^{\pici}_{\tau_1} (t) }
\end{gather}
    % -
    % \beta
\begin{gather}
\label{sec: g2}
G_2 = 
    \frac{1}{N_2} 
    \displaystyle\sum_{j=1}^{N_2} 
        \displaystyle\sum_{t=0}^{T_2}
        \nabla_\theta \log{\pim (a_{t2}^{(j)} \mid s_{t2}^{(j)} )} 
         \Tilde{A}^{\pim}_{\tau_2} (t) 
\end{gather}

where $c^{(i)}$ identifies the chosen expert of the trajectory., and $\Tilde{A}^{\pici}_{\tau_1} (t) = R_{\tau_1} (t) - \Tilde{V}^{\pici} (s_t)$ and $\Tilde{A}^{\pim}_{\tau_2} (t) = -\beta R_{\tau_2} (t) - \Tilde{V}^{\pim} (s_t)$ are the modified advantage functions. The $-\beta$ that is in the advantage in $G_2$ optimizes \textit{against} the performance of the observed policy $\pim$.

The gradient $G_1$ for $\pic$ is straightforward. 
However, to estimate the gradient $G_2$ for $\pim$ which is an fictitious policy, we sample from it by first re-sampling the context of the expert at each state, and then sampling an action from the context. 
The back-propagation occurs to $\pici (a \mid s)$ for the context sampled at each state. Practical implementation details can be found in \ref{sec: appendEst}.
The intuition is as follow. 
While sampling $\pim$, if a selected action causes {\textit{high}} return, we should {\textit{decrease}} the probability, which lowers the expected reward of $\pim$. 
Combined, the two gradients will cause the PE to select actions that both achieves high reward, and are detrimental to the observer.

% The $\nabla_\theta \log{\pim (a | s)}$ can be viewed as a mixture of experts \citet{mixexp}, where the gating network is fixed as $p(c | s)$, rather than being learnt.
Equations \ref{sec: returns} and \ref{sec: pgupdate} formulate our PG approach of APE, which is summarized in Algorithm \ref{alg:CAPE}. 

% \begin{wrapfigure}{R}{0.52\textwidth}
\begin{algorithm}[H]
\label{alg:CAPE}
\caption{PG-APE}
\begin{algorithmic}[1]
\Require{$\theta$, $\phi$, $\mathcal{M}$, $\beta$}
\State \textbf{for} each iteration \textbf{do}:  
    \Indent 
    \State Generate trajectories ${\tau}_1$ with $\pic$ from $\mathcal{M}$ for Equation \ref{sec: g1}
    \State Generate trajectories ${\tau}_2$ with $\pim$ from $\mathcal{M}$ for Equation \ref{sec: g2}
    
    % \State \textbf{for} each update \textbf{do}: \Indent
        \State Calculate Equation \ref{sec: pgupdate} to perform a gradient update on the PE $\theta \leftarrow \theta + \alpha_\theta \hat{\nabla}_\theta J_{\tau_1, \tau_2}(\theta)$
        \State Update the value function $\phi \leftarrow \phi - 
        \alpha_\phi \hat{\nabla}_\phi J_{\tau_1, \tau_2} (\phi)
        $ as determined by Equation \ref{sec: returns}.
    % \EndIndent \State \textbf{end for}
    
    \EndIndent
\State \textbf{end for}
\end{algorithmic}
\end{algorithm}
% \end{wrapfigure}

\section{Experiments}
\label{sec:result}

We perform experiments on a navigation task, where the objective is to reach a goal state as fast as possible.
The purpose is to illustrate that an APE can cause the cloned policy to take significantly longer to reach the goal state. 
We do so by first training a PE and behaviour cloning it. We then compare the performance of the PE to that of the clone.
% \ST{explain why this is the best setting to demonstrate the idea: POINTS: discrete, n=small, space no small. but all this wlog... }
We use a discrete environment to best demonstrate the validity of the equation. This is because all discrete policies can be parameterized, which is not true in continuous, where typically Gaussian parameterization is used. 
As such, continuous environments would have to make assumptions about how both the PE and the cloner parameterizes policies, as well as tackle problems of distributional drift, which we would like to avoid. 
However, with these assumptions, our setting can extend to the continuous domain.
In our experiments, we use a $10 \times 10$ grid-world environment as our main testbed. This is to have large enough expression that would not be found in smaller grids, while still small enough to visualize the behaviour of the APE. 
The discrete actions will show precisely how the experts can be jointly adversarial.

Using gridworld allows for precise expected return estimates. 
In an environment where there is no computable analytical solution for the returns, approximation error can accumulate through estimating the returns of both the trained PE and the clone. 
This noise would only increase in continuous state space, where the returns of $\pi_o$ may not be tractable to estimate due to issues such as distributional drift (\citet{dagger, BCLimitations, causal_imitation_learning}).

Our results answer the following questions.
How much optimality is compromised? 
How useless can we make the cloned policy?
Is it possible to use non APE to prevent behaviour cloning? 
% \footnote{\url{sites.google.com/view/prevent-imitation-learning}}

\subsection{Training}
\label{sec:navigation}

\begin{figure}[t]
    \captionsetup{justification=centering}
      \centering
      \includegraphics[scale=0.165]{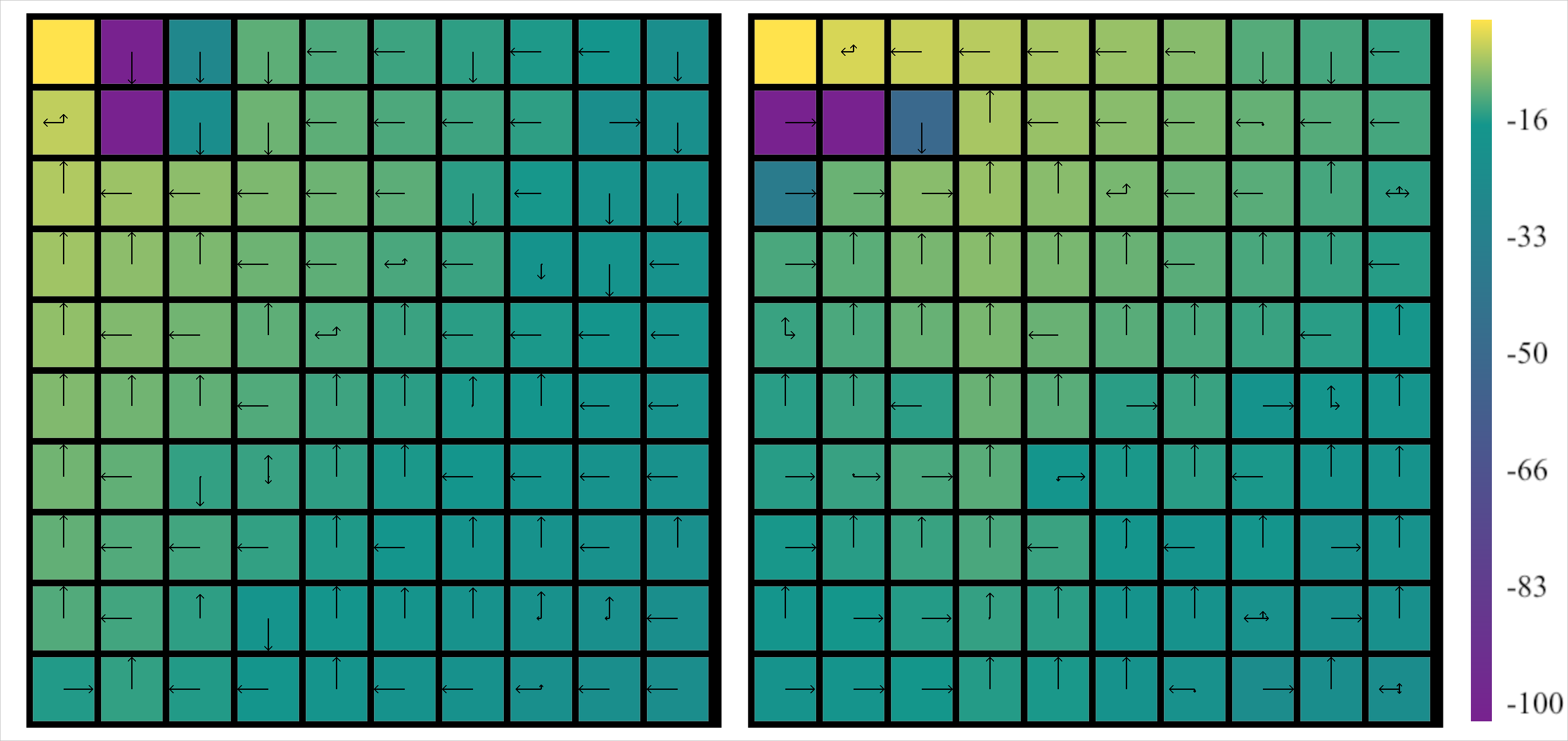}
      \captionof{figure}{\textbf{Visualization of APE for 2 experts}. We set $\beta = 0.6$.
      Arrows indicate action probabilities, and the colour scale represents the hitting time.
      Yellow indicates expected reward of 0, while
      purple indicates expected reward of $-100$, which is the maximum episode length. 
      The top left corner is the goal state, and the adjacent states that are purple are an example of how APE is adversarial to cloning, as those states will cause the cloned policy to suffer larger losses.}
      \label{fig:contexted}
\end{figure}

    Even though our method can compute a policy ensemble with any finite number of experts, we chose to visualize a solution with 2 experts, which is sufficient to reveal the essential properties of the method. Specifically, we train $n=2$ tabular experts with PG-APE. 
    Our code is written in Tensorflow (\citet{Tensorflow}). Training details and hyper-parameters are in Section \ref{sec:hyperparams} of the Appendix.

\subsection{Environment}

    The basic environment is a $10 \times 10$ grid, with the goal state at the top left corner.
    The agent spawns in a random non-goal state, and incurs a reward of $-1$ for each time-step until it reaches the goal. 
    At the goal state, the agent no longer receives a loss and terminates the episode.
    The agent is allowed five actions, $\mathcal{A} =$ \{ \textit{Up, Down, Left, Right, Stay} \}. 
    Moving into the wall is equivalent to executing a \textit{Stay} action.
    We choose this reward function for the benefit of having a clear representation of the notion of "good enough", which is reflected in how long it takes to reach the goal state. 
    Having such representation exemplifies how the APE can prevent an observer from cloning a good policy. 

\subsection{Visualization}
    
    Figure \ref{fig:contexted} shows an example of a PE that is trained for the basic gridworld environment. 
    Figure \ref{fig:cloned} shows the corresponding cloned policy, as well as a comparison to an optimal policy. 
    The colour scale represents the expected return of starting at a given state. 
    
    In the case of an optimal policy ($\beta=0$), actions are taken to take the agent to the goal state as fast as possible. 
    However, when $\beta > 0$, such a solution is no longer the optimum.
        Similar to $\beta=0$, the experts would like to maximize the expected reward, and reach the goal state. 
    However, to minimize the reward of the observed policy, the two expert policies must jointly learn to increase the number of steps needed for $\pim$ to reach the goal state.
    The expert policies must use adversarial behaviour while reaching the goal state, such as taking intelligent detours or \textit{Stay} in the same state, which are learned to hinder $\pim$ as much as possible.
    These learnt behaviours cause the cloned policy to take a drastically longer time to reach the goal.  
    For example, note the two purple squares at the top-left near the goal, which indicates that the experts understand that they should not move to prevent the observer from attaining reward. Even though these sub-optimal decisions are made, on expectation, the experts are "not bad" and achieve an average of $-15.27$ reward.
    
    % Reported in Table \ref{table:betaComp} are the mean and standard deviations of the hitting times for three different values of $\beta$, averaging across $10$ different seeds. 
    % As $\beta$ increases, the CPE becomes more and more adversarial, making incremental sacrifices in performance of $\pic$ 
    % for dramatic decreases in performance of $\pim$. 

\subsection{Baselines}

\begin{wrapfigure}{R}{0.52\textwidth}
    % \begin{subfigure}{.47\textwidth}
    %     \includegraphics[scale=0.33]{pics/optim.png}
    %   \label{fig:optimal}
    % \end{subfigure}
    \begin{subfigure}{.47\textwidth}
      \includegraphics[scale=0.165]{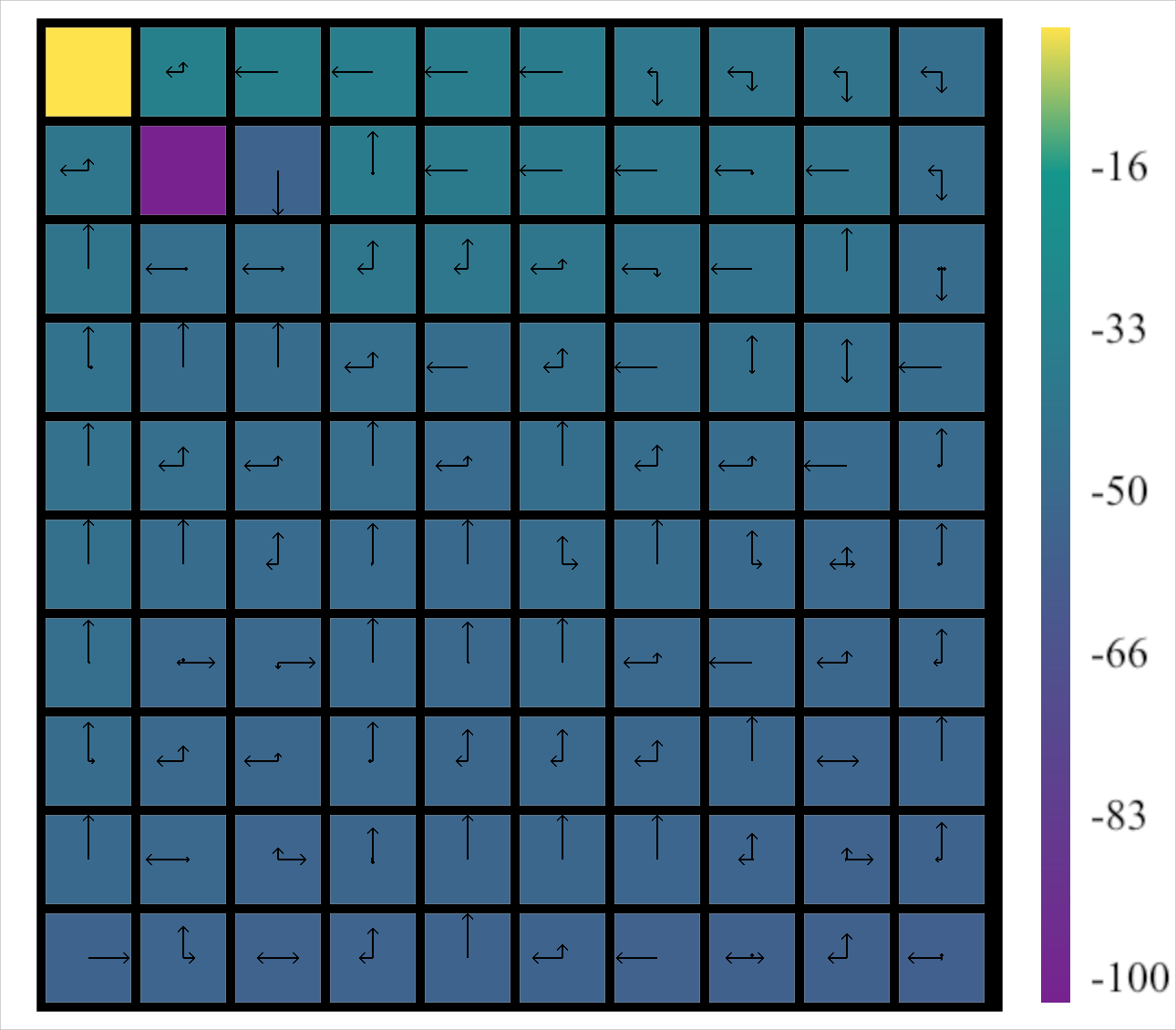}
      \label{fig:optimal}
    \end{subfigure}
    \caption{\textbf{Visualization of the cloned APE}. The policy obtained from cloning the APE trained has average expected reward of $-45.18$, while the optimal policy has an average expected reward of $-9$, which is over a $5\times$ increase.}
    \label{fig:cloned}
\end{wrapfigure}

We use behaviour cloning to clone our PG-APE trained policies. 
To support our claims of preventing IL even in the horizon of infinite data, we collect a million timesteps of the trained PE in the environment. 
Further details of behaviour cloning are in the appendix.
% We answer here that our trained policies can indeed prevent behaviour cloning.
Shown in Figure \ref{fig:cloned} is an optimal policy, and the resulting cloned policy from Section \ref{sec:navigation}.

We evaluate against other PE, to show that preventing against behaviour cloning is non-trivial.
We use several baselines.
We first test policies that have approximately the same return as our ensemble by training PE with vanilla PG, and halting early rather than running until convergence.
In the Near-Optimal case, we ran until the PE had expected returns that matched the average achieved by our method.
Conversely, "Random" policies are used as a comparison to show that it is possible to cause the cloned policy to do poorly, but the tradeoff is that the PE itself cannot perform well, which is undesirable. 
These policies are also policies trained with PG, except they are stopped much earlier, when their clones matches the expected returns of our PG-APE.
For each PG-APE, we use $n=2$ different tabular policies treated as an ensemble, which we then clone, and average across $5$ seeds. 
For the baselines, we hand-pick the policies, and thus only use $3$ different policies.

\newcolumntype{d}{c{c}{c}{c}}
\newcolumntype{C}{>{\centering}p}
\begin{table}[htp]
\centering
\begin{center}
\begin{tabular}{p{1.35in}dd}
\toprule
\multicolumn{1}{C{0.3in}}{} & 
\multicolumn{1}{C{1.0in}}{PE Returns} & 
\multicolumn{1}{C{1.0in}}{Clone Returns} &
\multicolumn{1}{C{1.2in}}{Returns Difference} \\
\midrule
PG-APE & -16.24 $\pm$ 1.20 & -44.27 $\pm$ 1.07 & \textbf{-28.03}\\
Near-Optimal PE & -16.74 $\pm$ 1.32 & -16.67 $\pm$ 1.31 & \texttt{+}0.07\\
Random Policy & -44.59 $\pm$ 0.52 & -44.52 $\pm$ 0.77 & \texttt{+}0.07\\
\bottomrule
\end{tabular}
\newline
\newline
\caption{\textbf{Comparison of cloned PE}. Each policy has their Returns precisely calculated through their analytical solutions. The final column reports the difference between the PE and the Clone, which is only significant for our method.}
\label{table:PEComp}
\end{center}
\label{default}
\end{table}
As presented in Table \ref{table:PEComp}, all other PE have an insignificant difference (returns of the PE subtracted from returns of the cloned policy) between the performance of the PE and the cloned policy, except for our method. 
These empirical findings show that preventing behaviour cloning difficult, but possible using APE.

% \newcolumntype{d}{D{.}{.}{3.2}{.}}
% \newcolumntype{C}{>{\centering}p}
% \begin{table}[htp]
% \caption{Comparison of different values of $\beta$}    
% \label{table:betaComp}
% \centering
% \begin{center}
% \begin{tabular}{p{1.35in}dddd}
% \toprule
% \multicolumn{1}{C{1.35in}}{} & 
% \multicolumn{1}{C{.75in}}{0 (optimal)} & 
% \multicolumn{1}{C{.5in}}{0.1} &         
% \multicolumn{1}{C{.5in}}{0.2} &
% \multicolumn{1}{C{.5in}}{0.3} \\
% \midrule
% Policy Ensemble & -9.00 & -9.52 $\pm$ 0.22 & -10.06 $\pm$ 0.37 & -10.77 $\pm$ 0.68\\
% Cloned Policy & -9.00 & -41.22 $\pm$ 20.41 & -144.76 $\pm$ 55.68 & -280.00 $\pm$ 106.62 \\
% % Contexted Ensemble Hitting Time Std & 0 & 0.22 & 0.37 & 0.68\\
% % Context Marginalized Hitting Time Std & 0 & 20.41 & 55.86 & 106.62\\
% \bottomrule
% \end{tabular}
% \end{center}
% \caption{The reward for the trained PE, as well as the cloned policy using behaviour cloning. The PE are averaged across 3 seeds, and we clone each seed 3 times and report the average.}
% \end{table}

\section{ Discussion \& Future Work }\label{sec:dicussion}
 \textbf{Confidential Policies:} 
 There are promising research directions regarding the protection of policies, due to the many applications where confidentiality is crucial. 
As long as there is a model of the observer, our presented method provides a worst-case scenario of experts.

% \textbf{The Expressiveness of Observers:} 
% \label{sec: append obs}
In our work, we focused on the case where the observer does not use the current trajectory to determine their policy. 
Instead, it may be the case that the observer uses a sequential policy (one that depends on its previous states and/or actions), such as an RNN to determine the context of the current expert. 

Formally, the observer will no longer learn the policy formulated in Equation \ref{eqn:observe} that is solely dependent on the current state, 
but rather a policy that is dependent on the current trajectory: 
% $\pim(a \mid \tau_{1:t}) \defeq \sum_i p(\ci \mid \tau_{1:t} ) \pici (a \mid s)$.
\begin{align}
\label{eqn:observeRNN}
\pim ({a} \mid \tau_{1:t}) 
\defeq
\sum_i p(\ci \mid \tau_{1:t} ) \pici ( a \mid s)
\end{align}
We found in our preliminary results that using an RNN classifier which outputs $p(c | \tau_{1:t})$ simply ended up in with either optimal policies or crippled policies.
In both cases, there was a relatively minor difference in performance between the policy ensemble and the cloned policy.

Unsurprisingly, when the observer has access to a strong enough representation for their policy, then they should be able to imitate \textit{any policy}. In this case, the worst-case set of experts cannot do much to prevent the cloning. We believe that this is an exciting conclusion, and is grounds for future work.
% However, the corollary is that if the worst case can be cloned, then RNN methods of imitation learning can hold much power
 
    \textbf{Continuous:}
    Although our methods are evaluated in discrete state spaces, our approach can be generalized to continuous domains.
    
    The Monte Carlo sampling in Equation \ref{eqn:exp_MC} suggests that the use of continuous context may also be possible, given there is a strong enough function approximator to estimate the distribution of $c | s$. 
    We see this as an exciting direction for future work, to recover the full spectrum of possible adversarial policies under the constraint of Equation \ref{eqn:constrained}.

    \textbf{The Semantics of Reward:}
    Although the minimization in Equation \ref{eqn:constrained} implies a logical equivalence between the success of behaviour cloning to the reward the cloned policy can achieve, it may follow that this is not the case.
    It may be the case that useless is defined differently by the expected reward the cloned policy achieves on a different reward function $\Tilde{r}$.
    For example, a robot that is unpredictable should not be deployed with humans.
    Since the $r$ functions in Equation \ref{eqn:constrained} are disentangled,
    the reward function $r$ that is minimized in Equation \ref{eqn:constrained} can be engineered to fit any definition of uselessness. 
    
    We can modify the objective of APE by modifying Equations \ref{sec: returns} and \ref{sec: pgupdate} to use a different reward function $\Tilde{r}$ in the minimization, substituting $R(t)$ for $\Tilde{R}(t) = \sum_{t'=t}^{T-1} \gamma^{t'-t} \Tilde{r} (s_{t'}, a_{t'})$. 
    The rest of the derivation and algorithm remain the same. 
    
    We think this is an exciting direction, especially for learning all different possible representations of the worst-case experts.

\section{Conclusion}
\label{sec:conclusion}
We present APE as well as its mathematical formulation, and show that policy gradient, a basic RL algorithm can be used to optimize a policy ensemble that cannot be cloned.
We evaluated APE against baselines to show that adversarial behaviour is not feasible without our method.

This work identifies a novel yet crucial area in Reinforcement Learning, regarding the confidentiality of proprietary policies. 
The essence of our approach is that a policy ensemble can achieve high return for the policy owner, while providing an external observer with a guaranteed low reward, making proprietary ensemble useless to the observer. 

The formulation of our problem setup and the algorithm are very general. 
In this first work we demonstrate the solution in the deliberately chosen simple environments in order to better visualize the essence of our method. In our concurrent work we study thoroughly the application of our method in various domains, which is out of the scope of this introductory paper.  

\section{Acknowledgements}
This work was supported in part by NSF under grant NRI-\#1734633 and by Berkeley Deep Drive.

\bibliographystyle{corlabbrvnat}
\bibliography{draft}

\begin{thebibliography}{42}
\providecommand{\natexlab}[1]{#1}
\providecommand{\url}[1]{\texttt{#1}}
\expandafter\ifx\csname urlstyle\endcsname\relax
  \providecommand{\doi}[1]{doi: #1}\else
  \providecommand{\doi}{doi: \begingroup \urlstyle{rm}\Url}\fi

\bibitem[Laskey et~al.(2017)Laskey, Lee, Hsieh, Liaw, Mahler, Fox, and
  Goldberg]{DART}
M.~Laskey, J.~Lee, W.~Y. Hsieh, R.~Liaw, J.~Mahler, R.~Fox, and K.~Goldberg.
\newblock Iterative noise injection for scalable imitation learning.
\newblock \emph{CoRR}, abs/1703.09327, 2017.
\newblock URL \url{http://arxiv.org/abs/1703.09327}.

\bibitem[Finn et~al.(2017)Finn, Yu, Zhang, Abbeel, and Levine]{oneshotIL}
C.~Finn, T.~Yu, T.~Zhang, P.~Abbeel, and S.~Levine.
\newblock One-shot visual imitation learning via meta-learning.
\newblock \emph{CoRR}, abs/1709.04905, 2017.
\newblock URL \url{http://arxiv.org/abs/1709.04905}.

\bibitem[Codevilla et~al.(2019)Codevilla, Santana, L{\'{o}}pez, and
  Gaidon]{BCLimitations}
F.~Codevilla, E.~Santana, A.~M. L{\'{o}}pez, and A.~Gaidon.
\newblock Exploring the limitations of behavior cloning for autonomous driving.
\newblock \emph{CoRR}, abs/1904.08980, 2019.
\newblock URL \url{http://arxiv.org/abs/1904.08980}.

\bibitem[Codevilla et~al.(2017)Codevilla, M{\"{u}}ller, Dosovitskiy,
  L{\'{o}}pez, and Koltun]{end2endCondIL}
F.~Codevilla, M.~M{\"{u}}ller, A.~Dosovitskiy, A.~L{\'{o}}pez, and V.~Koltun.
\newblock End-to-end driving via conditional imitation learning.
\newblock \emph{CoRR}, abs/1710.02410, 2017.
\newblock URL \url{http://arxiv.org/abs/1710.02410}.

\bibitem[Pomerleau(1988)]{alvinn}
D.~Pomerleau.
\newblock Alvinn: An autonomous land vehicle in a neural network.
\newblock In \emph{NIPS}, 1988.

\bibitem[Bojarski et~al.(2016)Bojarski, Testa, Dworakowski, Firner, Flepp,
  Goyal, Jackel, Monfort, Muller, Zhang, Zhang, Zhao, and
  Zieba]{end2endSelfDriving}
M.~Bojarski, D.~D. Testa, D.~Dworakowski, B.~Firner, B.~Flepp, P.~Goyal, L.~D.
  Jackel, M.~Monfort, U.~Muller, J.~Zhang, X.~Zhang, J.~Zhao, and K.~Zieba.
\newblock End to end learning for self-driving cars.
\newblock \emph{CoRR}, abs/1604.07316, 2016.
\newblock URL \url{http://arxiv.org/abs/1604.07316}.

\bibitem[Al-Rubaie and Chang(2019)]{privateML}
M.~Al-Rubaie and J.~M. Chang.
\newblock Privacy-preserving machine learning: Threats and solutions.
\newblock \emph{IEEE Security \& Privacy}, 17\penalty0 (2):\penalty0 49--58,
  2019.

\bibitem[Papernot et~al.(2016)Papernot, Abadi, Úlfar Erlingsson, Goodfellow,
  and Talwar]{PATE}
N.~Papernot, M.~Abadi, Úlfar Erlingsson, I.~Goodfellow, and K.~Talwar.
\newblock Semi-supervised knowledge transfer for deep learning from private
  training data, 2016.

\bibitem[Ziebart et~al.(2008)Ziebart, Maas, Bagnell, and Dey]{maxentIRL}
B.~D. Ziebart, A.~Maas, J.~A. Bagnell, and A.~K. Dey.
\newblock Maximum entropy inverse reinforcement learning.
\newblock \emph{AAAI Conference on Artificial Intelligence}, 2008.

\bibitem[Levine(2018)]{controlasoptimalinference}
S.~Levine.
\newblock Reinforcement learning and control as probabilistic inference:
  Tutorial and review.
\newblock \emph{CoRR}, abs/1805.00909, 2018.
\newblock URL \url{http://arxiv.org/abs/1805.00909}.

\bibitem[Sutton et~al.(2000)Sutton, Mcallester, Singh, and Mansour]{PG}
R.~Sutton, D.~Mcallester, S.~Singh, and Y.~Mansour.
\newblock Policy gradient methods for reinforcement learning with function
  approximation.
\newblock \emph{Adv. Neural Inf. Process. Syst}, 12, 02 2000.

\bibitem[Schulman et~al.(2015)Schulman, Moritz, Levine, Jordan, and
  Abbeel]{gae}
J.~Schulman, P.~Moritz, S.~Levine, M.~Jordan, and P.~Abbeel.
\newblock High-dimensional continuous control using generalized advantage
  estimation.
\newblock 06 2015.

\bibitem[Schulman et~al.(2017)Schulman, Wolski, Dhariwal, Radford, and
  Klimov]{PPO}
J.~Schulman, F.~Wolski, P.~Dhariwal, A.~Radford, and O.~Klimov.
\newblock Proximal policy optimization algorithms.
\newblock \emph{CoRR}, abs/1707.06347, 2017.
\newblock URL \url{http://arxiv.org/abs/1707.06347}.

\bibitem[Widrow and W.~Smith(1964)]{ILOriginal}
B.~Widrow and F.~W.~Smith.
\newblock Pattern recognizing control systems.
\newblock \emph{Computer Inf. Sci. (COINS) Proc.}, 01 1964.

\bibitem[Abbeel and Ng(2004)]{IRL}
P.~Abbeel and A.~Y. Ng.
\newblock Apprenticeship learning via inverse reinforcement learning.
\newblock In \emph{ICML}, 2004.

\bibitem[Ng and Russell(2000)]{algoIRL}
A.~Y. Ng and S.~J. Russell.
\newblock Algorithms for inverse reinforcement learning.
\newblock In \emph{Proceedings of the Seventeenth International Conference on
  Machine Learning}, ICML '00, pages 663--670, San Francisco, CA, USA, 2000.
  Morgan Kaufmann Publishers Inc.
\newblock ISBN 1-55860-707-2.
\newblock URL \url{http://dl.acm.org/citation.cfm?id=645529.657801}.

\bibitem[Levine et~al.(2011)Levine, Popovic, and Koltun]{nonlinear}
S.~Levine, Z.~Popovic, and V.~Koltun.
\newblock Nonlinear inverse reinforcement learning with gaussian processes.
\newblock 12 2011.

\bibitem[Ho and Ermon(2016)]{GAIL}
J.~Ho and S.~Ermon.
\newblock Generative adversarial imitation learning.
\newblock \emph{CoRR}, abs/1606.03476, 2016.
\newblock URL \url{http://arxiv.org/abs/1606.03476}.

\bibitem[Peng et~al.(2018)Peng, Kanazawa, Toyer, Abbeel, and Levine]{VDB}
X.~B. Peng, A.~Kanazawa, S.~Toyer, P.~Abbeel, and S.~Levine.
\newblock Variational discriminator bottleneck: Improving imitation learning,
  inverse rl, and gans by constraining information flow.
\newblock \emph{CoRR}, abs/1810.00821, 2018.
\newblock URL \url{http://arxiv.org/abs/1810.00821}.

\bibitem[Lin et~al.(2017)Lin, Hong, Liao, Shih, Liu, and Sun]{DRLAttack}
Y.~Lin, Z.~Hong, Y.~Liao, M.~Shih, M.~Liu, and M.~Sun.
\newblock Tactics of adversarial attack on deep reinforcement learning agents.
\newblock \emph{CoRR}, abs/1703.06748, 2017.
\newblock URL \url{http://arxiv.org/abs/1703.06748}.

\bibitem[Behzadan and Munir(2017)]{vulnerablePolicies}
V.~Behzadan and A.~Munir.
\newblock Vulnerability of deep reinforcement learning to policy induction
  attacks.
\newblock In \emph{International Conference on Machine Learning and Data Mining
  in Pattern Recognition}, pages 262--275. Springer, 2017.

\bibitem[Ma et~al.(2019)Ma, Zhang, Sun, and Zhu]{poison}
Y.~Ma, X.~Zhang, W.~Sun, and X.~Zhu.
\newblock Policy poisoning in batch reinforcement learning and control.
\newblock \emph{arXiv preprint arXiv:1910.05821}, 2019.

\bibitem[Gleave et~al.(2019)Gleave, Dennis, Kant, Wild, Levine, and
  Russell]{gleaveAttackRL}
A.~Gleave, M.~Dennis, N.~Kant, C.~Wild, S.~Levine, and S.~Russell.
\newblock Adversarial policies: Attacking deep reinforcement learning.
\newblock \emph{arXiv preprint arXiv:1905.10615}, 2019.

\bibitem[Wang and Hegde(2019)]{privateQ}
B.~Wang and N.~Hegde.
\newblock Private q-learning with functional noise in continuous spaces.
\newblock \emph{arXiv preprint arXiv:1901.10634}, 2019.

\bibitem[Balle et~al.(2016)Balle, Gomrokchi, and Precup]{privatePolicy}
B.~Balle, M.~Gomrokchi, and D.~Precup.
\newblock Differentially private policy evaluation.
\newblock In \emph{International Conference on Machine Learning}, pages
  2130--2138, 2016.

\bibitem[Li et~al.(2017)Li, Song, and Ermon]{infogail}
Y.~Li, J.~Song, and S.~Ermon.
\newblock Inferring the latent structure of human decision-making from raw
  visual inputs.
\newblock \emph{CoRR}, abs/1703.08840, 2017.
\newblock URL \url{http://arxiv.org/abs/1703.08840}.

\bibitem[Brown et~al.(2019)Brown, Cui, and Niekum]{RiskAwareIRL}
D.~S. Brown, Y.~Cui, and S.~Niekum.
\newblock Risk-aware active inverse reinforcement learning.
\newblock \emph{CoRR}, abs/1901.02161, 2019.
\newblock URL \url{http://arxiv.org/abs/1901.02161}.

\bibitem[de~Haan et~al.(2019)de~Haan, Jayaraman, and
  Levine]{causal_imitation_learning}
P.~de~Haan, D.~Jayaraman, and S.~Levine.
\newblock Causal confusion in imitation learning.
\newblock \emph{CoRR}, abs/1905.11979, 2019.
\newblock URL \url{http://arxiv.org/abs/1905.11979}.

\bibitem[Hristov et~al.(2018)Hristov, Lascarides, and
  Ramamoorthy]{latentFromDemo}
Y.~Hristov, A.~Lascarides, and S.~Ramamoorthy.
\newblock Interpretable latent spaces for learning from demonstration.
\newblock \emph{CoRR}, abs/1807.06583, 2018.
\newblock URL \url{http://arxiv.org/abs/1807.06583}.

\bibitem[Achiam et~al.(2018)Achiam, Edwards, Amodei, and Abbeel]{valor}
J.~Achiam, H.~Edwards, D.~Amodei, and P.~Abbeel.
\newblock Variational option discovery algorithms.
\newblock \emph{CoRR}, abs/1807.10299, 2018.
\newblock URL \url{http://arxiv.org/abs/1807.10299}.

\bibitem[Eysenbach et~al.(2018)Eysenbach, Gupta, Ibarz, and Levine]{diayn}
B.~Eysenbach, A.~Gupta, J.~Ibarz, and S.~Levine.
\newblock Diversity is all you need: Learning skills without a reward function.
\newblock \emph{CoRR}, abs/1802.06070, 2018.
\newblock URL \url{http://arxiv.org/abs/1802.06070}.

\bibitem[Sharma et~al.(2019)Sharma, Gu, Levine, Kumar, and Hausman]{DADS}
A.~Sharma, S.~Gu, S.~Levine, V.~Kumar, and K.~Hausman.
\newblock Dynamics-aware unsupervised discovery of skills.
\newblock \emph{CoRR}, abs/1907.01657, 2019.
\newblock URL \url{http://arxiv.org/abs/1907.01657}.

\bibitem[Jacobs et~al.(1991)Jacobs, Jordan, Nowlan, Hinton, et~al.]{mixexp}
R.~A. Jacobs, M.~I. Jordan, S.~J. Nowlan, G.~E. Hinton, et~al.
\newblock Adaptive mixtures of local experts.
\newblock \emph{Neural computation}, 3\penalty0 (1):\penalty0 79--87, 1991.

\bibitem[Henderson et~al.(2018)Henderson, Chang, Bacon, Meger, Pineau, and
  Precup]{optiongan}
P.~Henderson, W.-D. Chang, P.-L. Bacon, D.~Meger, J.~Pineau, and D.~Precup.
\newblock Optiongan: Learning joint reward-policy options using generative
  adversarial inverse reinforcement learning.
\newblock In \emph{Thirty-Second AAAI Conference on Artificial Intelligence},
  2018.

\bibitem[Zhang et~al.(2019)Zhang, Yu, and Turk]{novel}
Y.~Zhang, W.~Yu, and G.~Turk.
\newblock Learning novel policies for tasks.
\newblock \emph{CoRR}, abs/1905.05252, 2019.
\newblock URL \url{http://arxiv.org/abs/1905.05252}.

\bibitem[Goyal et~al.(2019)Goyal, Islam, Strouse, Ahmed, Larochelle, Botvinick,
  Levine, and Bengio]{InfoBot}
A.~Goyal, R.~Islam, D.~Strouse, Z.~Ahmed, H.~Larochelle, M.~Botvinick,
  S.~Levine, and Y.~Bengio.
\newblock Transfer and exploration via the information bottleneck.
\newblock In \emph{International Conference on Learning Representations}, 2019.
\newblock URL \url{https://openreview.net/forum?id=rJg8yhAqKm}.

\bibitem[Strouse et~al.(2018)Strouse, Kleiman{-}Weiner, Tenenbaum, Botvinick,
  and Schwab]{LSH}
D.~Strouse, M.~Kleiman{-}Weiner, J.~Tenenbaum, M.~Botvinick, and D.~J. Schwab.
\newblock Learning to share and hide intentions using information
  regularization.
\newblock \emph{CoRR}, abs/1808.02093, 2018.
\newblock URL \url{http://arxiv.org/abs/1808.02093}.

\bibitem[Ross et~al.(2010)Ross, Gordon, and Bagnell]{dagger}
S.~Ross, G.~J. Gordon, and J.~A. Bagnell.
\newblock No-regret reductions for imitation learning and structured
  prediction.
\newblock \emph{CoRR}, abs/1011.0686, 2010.
\newblock URL \url{http://arxiv.org/abs/1011.0686}.

\bibitem[Abadi et~al.(2016)Abadi, Barham, Chen, Chen, Davis, Dean, Devin,
  Ghemawat, Irving, Isard, Kudlur, Levenberg, Monga, Moore, Murray, Steiner,
  Tucker, Vasudevan, Warden, Wicke, Yu, and Zhang]{Tensorflow}
M.~Abadi, P.~Barham, J.~Chen, Z.~Chen, A.~Davis, J.~Dean, M.~Devin,
  S.~Ghemawat, G.~Irving, M.~Isard, M.~Kudlur, J.~Levenberg, R.~Monga,
  S.~Moore, D.~G. Murray, B.~Steiner, P.~A. Tucker, V.~Vasudevan, P.~Warden,
  M.~Wicke, Y.~Yu, and X.~Zhang.
\newblock Tensorflow: {A} system for large-scale machine learning.
\newblock \emph{CoRR}, abs/1605.08695, 2016.
\newblock URL \url{http://arxiv.org/abs/1605.08695}.

\bibitem[Mnih et~al.(2016)Mnih, Badia, Mirza, Graves, Lillicrap, Harley,
  Silver, and Kavukcuoglu]{asynch}
V.~Mnih, A.~P. Badia, M.~Mirza, A.~Graves, T.~P. Lillicrap, T.~Harley,
  D.~Silver, and K.~Kavukcuoglu.
\newblock Asynchronous methods for deep reinforcement learning.
\newblock \emph{CoRR}, abs/1602.01783, 2016.
\newblock URL \url{http://arxiv.org/abs/1602.01783}.

\bibitem[Kingma and Ba(2015)]{AdaM}
D.~P. Kingma and J.~Ba.
\newblock Adam: A method for stochastic optimization.
\newblock \emph{CoRR}, abs/1412.6980, 2015.

\bibitem[Cho et~al.(2014)Cho, van Merrienboer, G{\"{u}}l{\c{c}}ehre, Bougares,
  Schwenk, and Bengio]{GRU}
K.~Cho, B.~van Merrienboer, {\c{C}}.~G{\"{u}}l{\c{c}}ehre, F.~Bougares,
  H.~Schwenk, and Y.~Bengio.
\newblock Learning phrase representations using {RNN} encoder-decoder for
  statistical machine translation.
\newblock \emph{CoRR}, abs/1406.1078, 2014.
\newblock URL \url{http://arxiv.org/abs/1406.1078}.

\end{thebibliography}

\newpage
\appendix
\section{Appendix}

\subsection{Training Details \& Hyperparameters}
\label{sec:hyperparams}

For our training, we set $\alpha_\theta = 0.05$, and the value weight to be $0.5$, use annealed entropy regularization (\citet{asynch}) from $5 \text{e}-1$ to $5 \text{e}-3$ and set the discount factor $\gamma=0.99$.
Due to the contrasting gradients experienced, large batch sizes are used.
In our experiments, we take 1 gradient update of AdaM (\citet{AdaM}) per batch of 4096 (containing multiple trajectories), and trained for $3e6$ timesteps.

To estimate $p(c|s)$ in Equation \ref{eqn:observe}, we use a replay buffer that keeps track of the previous $60$ contexts seen at each state.

Estimating the quantity in Equation \ref{eqn:observeRNN} requires memory, which we use a single GRU (\citet{GRU}) as done in \citet{LSH}, with the exception that only states are fed in as a one-hot. 
Due to our environment is deterministic, state sequences captures the action sequence information.
The single unit is then concatenated with the state, which feeds into a fully connected layer of 128, and then a soft-max, to produce the distribution $c | s$ over contexts.

For our behaviour cloning, we collect $1e6$ state action pairs, and train a tabular policy with $0.01$ learning rate on cross entropy softmax loss for $100$ epochs. 
The large amount of data and epochs is to ensure that we can recover $\pim$ with little to no variance. 

To solve the precise returns of the policies, we inject noise of $1e-9$, to ensure a hitting time always exists from each state. 
As well, we clip all the hitting times to $T = 100$.

\subsection{Estimating $\nabla_\theta \log \pim$}
\label{sec: appendEst}
It is not obvious how $\nabla_\theta \log \pim$ \textit{should} be estimated, since $\pim$ is never realized until the policy is cloned. Literally, it is a virtual policy.

Equation \ref{eqn:observe} offers a straightforward method to back-propagate, similar to that of the Mixture of Experts model (\citet{mixexp}), except using an estimate of $c | s$ instead of a gating network.

However, we can also rewrite Equation \ref{eqn:observe} as $\sum_i p(\ci | s) \pici ( a \mid s)  
=
\E_{c \sim p(c | s)} [ \pici ( a \mid s)] $, which results in the gradient update being:

\begin{align}
\label{eqn:exp_MC}
\nabla_\theta \log \pim (a | s) = \nabla_\theta \log
\E_{c \sim p(c | s)} [ \pici ( a \mid s)]
\end{align}

which suggests a method of Monte Carlo sampling the inner expectation with $1$ sampled context. 
Empirically, we use the Monte Carlo sampling method.

\end{document}